# MaTableGPT: GPT-based Table Data Extractor from Materials Science Literature

Gyeong Hoon Yi,[1,2,+] Jiwoo Choi,[1,2,+] Hyeongyun Song,[1,+] Olivia Miano,[3] Jaewoong Choi,[1] Kihoon Bang,[1] Byungju Lee,[1] Seok Su Sohn,[2] David Buttler,[3] Anna Hiszpanski,[3,*] Sang Soo Han,[1,*] Donghun Kim[1,*]

[1]Computational Science Research Center, Korea Institute of Science and Technology, Seoul 02792, Republic of Korea

[2]Department of Materials Science and Engineering, Korea University, Seoul 02841, Republic of Korea

[3]Materials Science Division, Lawrence Livermore National Laboratory, Livermore, CA 94550, USA

[+]These authors contributed equally.

[*]Correspondence to: Donghun Kim (donghun@kist.re.kr), Sang Soo Han (sangsoo@kist.re.kr), Anna Hiszpanski (hiszpanski2@llnl.gov)

# Abstract

Efficiently extracting data from tables in the scientific literature is pivotal for building large-scale databases. However, the tables reported in materials science papers exist in highly diverse forms; thus, rule-based extractions are an ineffective approach. To overcome this challenge, we present MaTableGPT, which is a GPT-based table data extractor from the materials science literature. MaTableGPT features key strategies of table data representation and table splitting for better GPT comprehension and filtering hallucinated information through follow-up questions. When applied to a vast volume of water splitting catalysis literature, MaTableGPT achieved an extraction accuracy (total F1 score) of up to 96.8%. Through comprehensive evaluations of the GPT usage cost, labeling cost, and extraction accuracy for the learning methods of zero-shot, few-shot and fine-tuning, we present a Pareto-front mapping where the few-shot learning method was found to be the most balanced solution owing to both its high extraction accuracy (total F1 score>95%) and low cost (GPT usage cost of 5.97 US dollars and labeling cost of 10 I/O paired examples). The statistical analyses conducted on the database generated by MaTableGPT revealed valuable insights into the distribution of the overpotential and elemental utilization across the reported catalysts in the water splitting literature.

## Introduction

The use of machine learning (ML) for data-driven material discovery is becoming increasingly important. Materials informatics[1,2,3,4] can leverage computational databases such as the Materials Project[5], Open Quantum Materials Database (OQMD)[6], Novel Materials Discovery (NOMAD)[7], and Open Catalyst 2022 (OC22)[8]. However, these databases have limitations because they do not directly integrate the data obtained from the actual experiments. In this regard, considering that millions of scientific studies are reported in natural language, leveraging natural language processing (NLP) technology to efficiently extract and process data from the scientific literature is considered a highly promising approach.

The scientific literature contains data in various formats, such as text, figures, and tables, and different approaches are required for extracting each format of the data. In the text domain, methods such as ChemDataExtractor[9,10] and named entity recognition (NER)[11,12] enable the efficient extraction and processing of data. Recent efforts involving vision-based methods for extracting data from graphs have been reported in the scientific literature[13,14,15,16]. Unfortunately, however, attempts to extract data from tables are relatively rare because tables in the scientific literature exist in highly diverse forms, and the diversity of information encapsulated in tables is also vast.

Nevertheless, tables in the scientific literature provide significant advantages over text and figures for several reasons. First, tables often include not only data from the respective study but also various data points used as references. For example, tables for catalyst performance in catalysis papers often include data on various catalysts beyond the subject catalyst that can also be extracted. Second, tables provide pre-categorized data. For example, a table on catalyst performance may contain various performance metrics, such as catalyst activity and stability. The extraction of all data from a table can provide different types of performance data on catalysts in a single process; this extraction is more efficient and straightforward than text-based extraction methods that require individually accessing and extracting each performance data point. Third, in text, after recognizing entities through NER, the relationships between each entity need to be defined to extract the relationships between entities[17]. However, in tables, entities already exist in relation to each other within the rows and columns, and no additional techniques are needed to extract the relationships between the entities. For these reasons, tables are an ideal target for effectively constructing large databases, and attempts to accurately extract entire data from diverse forms of tables in the literature are needed. Recently, large language models (LLMs) have shown impressive results across various tasks[18,19,20,21]. LLMs such as the GPT can generate realistic and consistent results even without explicit training[22]. ML-based NLP technology enables efficient extraction and processing of data from the scientific literature, contributing significantly to materials informatics and the advancement of materials science research[23]. Therefore, leveraging LLMs such as GPT to consistently extract data from tables of diverse forms is a rational approach[20].

Here, we report MaTableGPT, which is a GPT-based table data extractor from the materials science literature. MaTableGPT uniquely presents two key strategies, table data representation and table splitting, for better GPT comprehension and for filtering hallucinated information via follow-up questions. According to a comprehensive evaluation of the literature on water splitting catalysis, MaTableGPT achieved an impressive extraction score of 96.8%. By assessing the GPT usage cost, labeling cost, and extraction accuracy across various zero-shot, few-shot, and fine-tuning learning methods, we revealed a Pareto-front solution. Among these methods, the few-shot learning method emerged as the most balanced solution, providing both a high extraction score (>95%) and low cost (GPT usage cost of 5.97 US dollars and labeling cost of only 10 I/O paired examples). Additionally, statistical analyses of the database obtained from the water splitting catalysis literature by MaTableGPT revealed the distribution of overpotentials as well as the elemental utilization of reported catalysts.

## Results

**Workflow of MaTableGPT.** By applying MaTableGPT to the 2,406 tables in the 11,077 water splitting catalysis papers, we have a total of 47,670 catalytic performance data points for 12,122 catalyst materials. This database includes the catalyst name, reaction type, catalytic performance and corresponding measurement conditions, such as the overpotential measured in a specific electrolyte and applied voltage. MaTableGPT enabled a data extraction score of 96.8%. The overall workflow of MaTableGPT for this extraction is shown in **Fig. 1**.

First, a keyword-based search of publishers was conducted for paper collection. Term frequency-inverse document frequency (TF-IDF) was subsequently employed to classify the documents and eliminate the noisy papers: this resulted in the acquisition of a final set of 11,077 papers. Further details regarding content acquisition are described in the Methods section. Moreover, rules were applied to classify only tables related to the performance of catalysts. The tables for the data extraction exist in HTML format.

The table complexity in the materials science area, coupled with the presence of numerous unnecessary tags in HTML tables, poses significant challenges to data extraction. To address these issues, two strategies were implemented to create inputs that were easily understandable by the GPT model. The HTML tables were transformed into formats of either JavaScript Object Notation (JSON) and Tab-Separated Values (TSV) to remove unnecessary tags, and table splitting was performed to reduce the input length while lowering the possibility of cross-extraction. Using the newly created inputs, the potential of various learning methods, which include fine-tuned GPT models as well as zero-shot and few-shot learning methods with substantially lower labeling costs, was explored. Afterward, follow-up questions were asked to improve the data extraction accuracy by removing the hallucinated information without additional labeling, thereby improving the final data performance. The obtained data were then used to conduct several data mining studies.

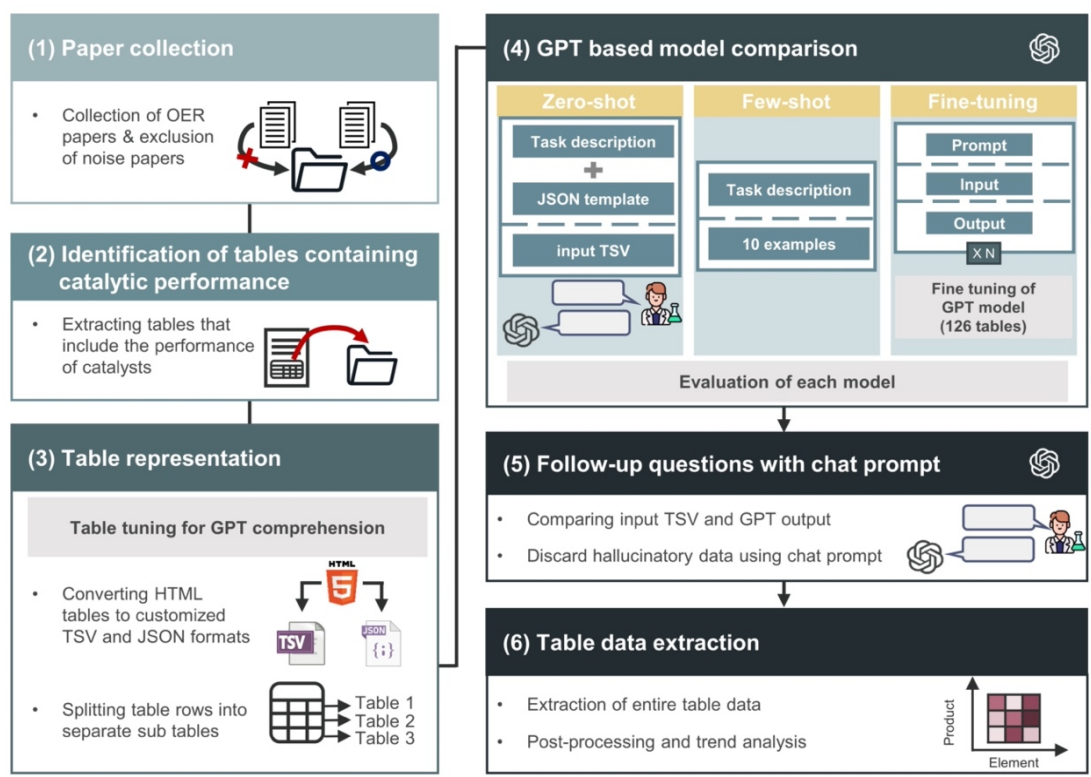

**Fig. 1 Overall workflow of MaTableGPT.** (**1**) Step 1: collection of papers relevant to oxygen evolution reaction (OER) and excluding noise papers. (**2**) Step 2: identification of tables containing the catalytic performance data. (**3**) Step 3: table data representation to aid GPT's comprehension. (**4**) Step 4: GPT training with zero-shot, few-shot, and fine-tuning approaches. (**5**) Step 5: follow-up questions to reduce hallucinatory data. (**6**) Step 6: building the database using the pretrained MaTableGPT model.

**Table data representation and table splitting.** In a collection of 11,077 papers, 4,905 tables were identified. Among these tables, the intention was to locate those tables detailing the performance of the catalysts. Upon categorizing the entire set of tables, distinctions could be made between the tables containing catalytic performance (e.g., overpotential, current density), tables containing calculated data (e.g., DFT calculations), tables containing characterization data (e.g., XRD), and others classified as noise. Rules based on the keyword analysis were established to allocate the tables into these four categories. Approximately 100 tables were placed into the noise category. The rule-based classification processes finally led to 2,406 tables in the category of tables containing catalytic performance, and these tables were used in the remaining ML studies. Due to a preference for experimental data, tables in the other categories were excluded mainly because they could not be processed through NLP techniques.

**Fig. 2** illustrates the appearance of tables in the materials science literature in various forms, with the four types of examples. **Fig. 2a** is an example of a table with merged cells. When the header rows are multiple and merged in different ways, the determination of header class a data point belongs to can be very difficult. Additionally, some tables are in HTML format where cells within the body play the role of headers, as shown in **Fig. 2b**. In some cases, as in **Fig. 2c**, information may be briefly summarized with a caption indexed within the table, while the necessary details are provided outside the table. Furthermore, tables exist in both conventional table formats (catalyst names in the leftmost column) and transposed formats (catalyst names in the topmost row), as shown in **Fig. 2d**. Due to these diverse formats, it is quite challenging to extract data from tables using rule-based methods; thus, more intelligent ways are needed to leverage LLMs such as GPT models.

Tables can also consist of various expressions even for the same term, which adds to the complexity. For instance, terms such as 'overpotential' can be expressed in different ways, such as 'overpotential at 10 mA/cm$^2$,' '$\eta$ at 10 mA/cm$^2$' or simply '$\eta_{10}$.' This variability is not limited to 'overpotential' but extends to other properties as well. While NER addresses these issues in the text, performing NER directly on HTML-formatted tables is impractical.

**Fig. 2 Examples of various formats of tables reported in the materials science literature.** (**a**) Table with a multilayered header of four rows and merged cells[24]. (**b**) Multi-header table with sub-headers named the hydrogen evolution reaction (HER) and OER, each with two rows detailing catalyst performance[25]. (**c**) Table including a caption and its index. The yellow frames outside the table explain the meaning of the caption index, which is denoted as the Greek letter or abbreviation[26]. (**d**) Transposed table with rows and columns reversed compared to standard tables with catalyst names written in the leftmost column[27].

To enable GPT to effectively comprehend these complex tables, we applied two strategies: table data representation and table splitting; each process is illustrated in **Figs. 3** and **4**. First, the process of table data representation is explained. HTML, as a markup language for web browsers, often contains a considerable amount of unnecessary information in its tables. These extraneous tags not only cause lengthy input but can also decrease the accuracy of the GPT models. To address these challenges, a reorganization of HTML into different formats was performed. Two custom formats were developed, each based on JSON and TSV. Due to its lightweight nature, JSON features a human-readable structure that is easy for humans to read and write and for machines to parse and generate. On the other hand, TSV emphasizes simplicity by storing tab-separated data in a straightforward text-based format. The top table in **Fig. 3** illustrates how the components of HTML tags are reflected in each customized JSON and customized TSV.

The conversion to the customized JSON format first involves identifying information regarding superscripts, subscripts, and captions and then transforming the HTML tags according to the rules outlined in the table in **Fig. 3a**. The modified HTML tags are then converted into a dataframe format using the Python library pandas. Subsequently, any merged rows and columns in the table represented as a dataframe are identified and split into individual cells to ensure that no cells remain merged. The values from the merged cells are copied and pasted into the corresponding empty cells. This approach results in both the complex header and merged body sections being transformed into a simpler table format. Finally, for each column, the header becomes the key, and the body becomes the value, which is then stored in JSON format. Additionally, the title and caption information from the HTML tags are extracted and included in the JSON; this completes the customized JSON representation. An example of this conversion process is shown in **Figs. 3b-3d**. JSON is widely known for its machine-friendly structure; thus, it is an effective choice for representing tables. However, its main drawback is that since headers are represented as the keys and the bodies as values on a per-column basis, the cases of multi-header tables with header in the body may lead to errors.

In the TSV format, the cells on the same line are typically separated by \t, and the lines are generally separated by \n. By adding simplified HTML tags to describe the table's title, table, caption, merged cells, superscript, and subscript, the details of the table were made understandable to the GPT model. An example of this conversion process is shown in **Figs. 3b**, **3c**, and **3e**. **Fig. 3f** shows an example of the final output (in JSON format) produced via GPT predictions with inputs in either customized JSON or TSV formats.

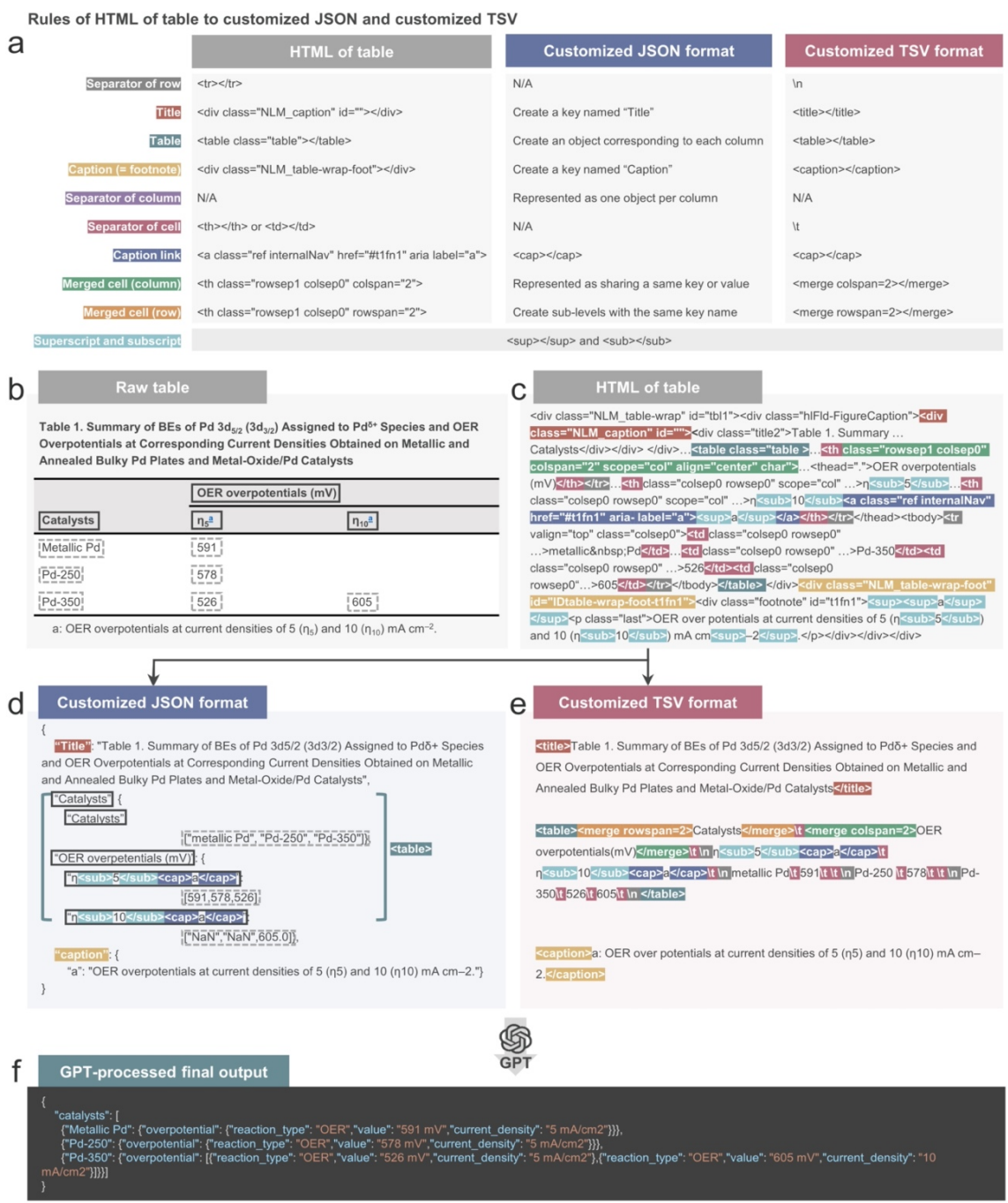

**Fig. 3 Table data representation from the HTML format to customized JSON or TSV formats for effective GPT comprehension.** (**a**) Conversion rules from the HTML tags to customized JSON or TSV formats. (**b-e**) Examples of a (b) raw table and its representations in (c) HTML, (c) customized JSON format, and (d) customized TSV format. (**f**) Example of the final outcome (in JSON) produced by the GPT predictions.

Through table data representation, we created an input structure that was easily understandable by the GPT model. However, both GPT-4 and GPT-3 have limitations in terms of the number of output tokens, with 4096 tokens for GPT-4-1106-preview and GPT-3.5-turbo-106. These limitations restrict the representation of all information from a large table consisting of dozens of rows. Additionally, when all data are extracted from a single table, cross-extraction can potentially occur. For instance, if the overpotential data of two different catalysts are interchanged during extraction, the extracted overpotential values, even if accurate, cannot be considered reliable data. To effectively address these issues, we conducted table splitting to alleviate concerns regarding token limitations in the GPT and to mitigate the possibility of cross-extraction, as illustrated in **Fig. 4**.

Here, a table is reorganized into multi-tables by dividing the table body into individual rows. In **Fig. 4a**, the splitting process for a typical table consisting of a title, header, body, and caption is shown. The table body is divided into rows, with each segment then attached to the other components to produce multiple tables. However, the variety of table formats extends beyond the typical formats, and not all tables conform to this standard structure. **Fig. 4b** presents two examples of atypical tables: a multi-header table and a table with headers within the body. The case of the multi-header table includes sub-headers within the table body, and these sub-headers are classified and placed directly below the main header during the splitting process. For tables where headers are found within the body, these headers need to be identified and appropriately positioned above the corresponding body, ensuring clarity and coherence in the restructured tables. To facilitate this, rules for dividing atypical tables were developed after extensively examining the various table formats and are provided in codes in the **Code Availability** section.

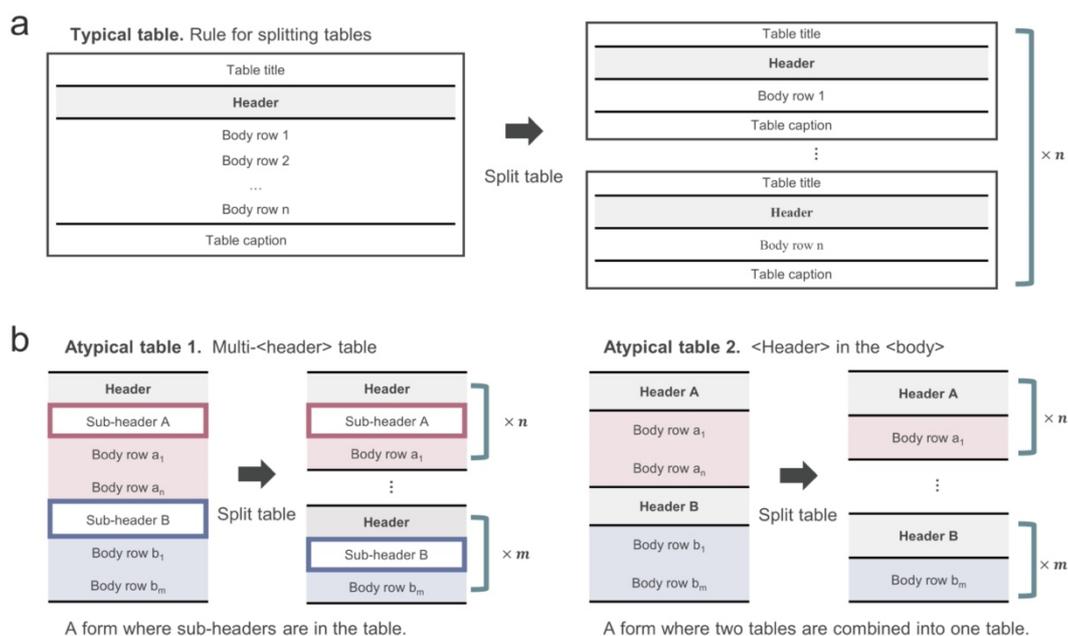

**Fig. 4 Processes and rationale behind table splitting.** (**a**) Table splitting rules for a typical table. (**b**) Examples of atypical tables with header complexities and rules for table splitting therein.

**Training GPT models.** Using the provided input obtained through the above two strategies (table data representation and table splitting), we now perform GPT modeling based on fine-tuning, few-shot learning, and zero-shot learning, as illustrated in **Fig. 5**. Recent studies have demonstrated reliable performance by pretraining on large quantities of data, independent of the downstream task, followed by task-specific fine-tuning[28,29]. However, considering the substantial labeling costs associated with fine-tuning and the strengths of generative models such as GPT in few-shot and zero-shot learning, exploring alternative learning methods is crucial. In this regard, we employ various approaches of fine-tuning, few-shot and zero-shot learning methods. Additionally, we propose a method using follow-up questions to effectively mitigate hallucinated information in the GPT outputs.

First, for the fine-tuning method, 126 input–output pairs of tables were created for training and testing. The dataset for fine-tuning consists of three parts. The first part serves as a prompt for the task and describes the actions required of the GPT model. The second part consists of an input, and this input is prepared in the original HTML, the customized JSON, and the customized TSV formats of a table. The third part involves an output, and this output represents the desired result obtained through the GPT prediction process and is formatted in JSON.

Second, for few-shot learning, ten diverse and complex table examples sourced exclusively from the material domain are curated to form input–output pairs. These pairs represent a spectrum of intricate table structures, consisting of complex tables, as illustrated in **Fig. 2**, and conventional table examples. The selected examples are input into the GPT model along with carefully crafted task descriptions. These task descriptions provide a list of performance metrics and the desired property information to be extracted. Detailed task descriptions can be found in Supplementary Note 1. Subsequently, without the necessity of additional training, data extraction is directly applied to the designated table input. This departure from fine-tuning has a significant advantage in reducing labeling costs.

Third, zero-shot learning involves providing a JSON TEMPLATE for filling data alongside a description of the task. The process subsequently proceeds by sequentially asking for the required information while providing the input representation. JSON TEMPLATE is structured hierarchically, with the catalyst name, performance, and performance-related attributes (or properties). Accordingly, the process begins by inquiring about the presence of a catalyst, followed by querying information regarding performance and its associated properties. The data present in the actual input are then entered into the JSON TEMPLATE. Finally, any remaining unfilled property keys are removed. Since no labeling cost is needed, effective data extraction can be achieved across domains simply by modifying the template without being constrained by domain-specific limitations. Examples of zero-shot learning can be found in Supplementary Note 2.

The output data obtained through fine-tuning, few-shot, and zero-shot learning methods may contain hallucinated information. Since the outputs derived from each model adhere to the structure of the JSON TEMPLATE, a discrepancy in catalyst names, even with correctly extracted performance and property information, causes the entire dataset to be incorrect. Since the accuracy of the higher-level hierarchy critically impacts the data quality, follow-up questions are employed to verify the correctness of the data extraction at each level. When inaccuracies are detected, the key is removed to reduce the amount of hallucinated information. The questions are straightforward: both the input TSV and GPT output are provided, and the process regarding the catalyst, performance, and performance-related properties that are present in the GPT output but not in the input TSV are examined. If any discrepancies are found, the corresponding data are removed.

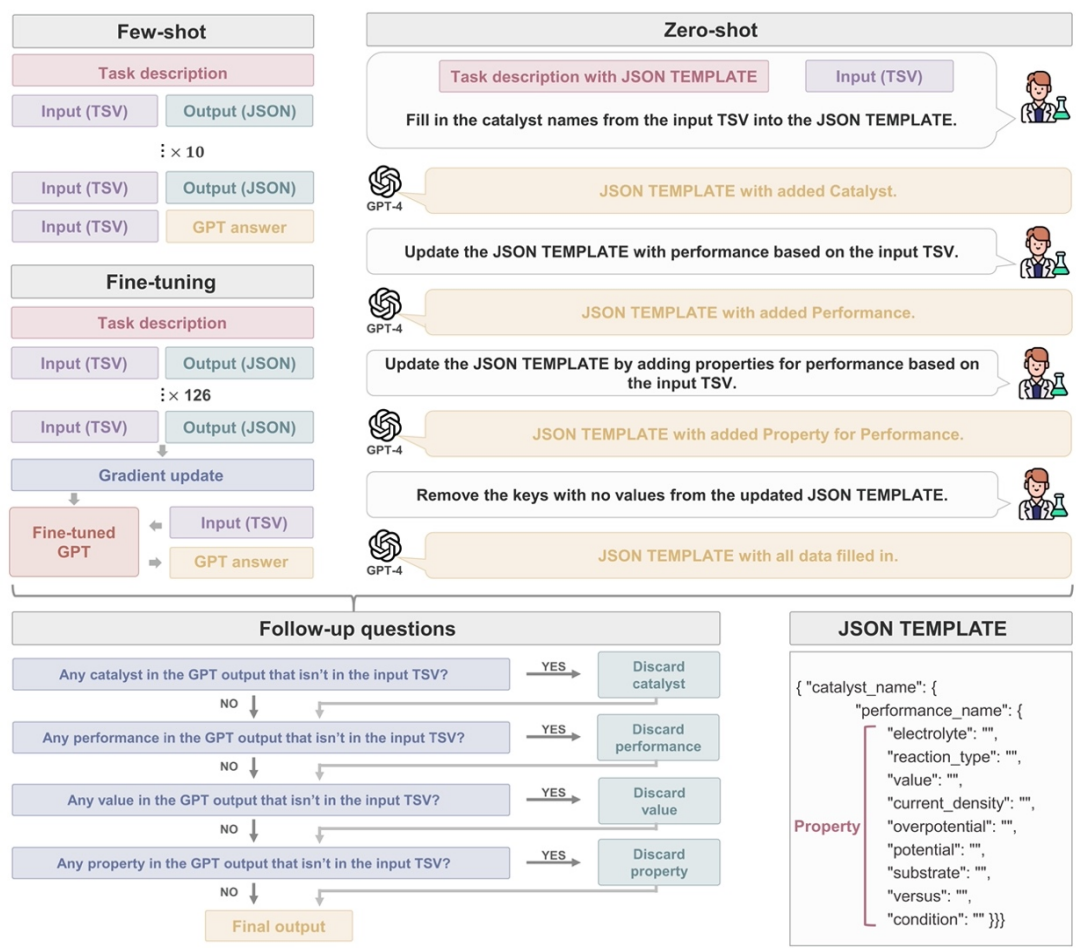

**Fig. 5 Schematic showing the GPT modeling process based on fine-tuning, few-shot, and zero-shot learning schemes.** The extracted databases obtained by the three learning methods are in JSON template format, and each database is subjected to conversational follow-up questions to minimize hallucinations. In this schematic, customized TSV, instead of customized JSON, is chosen as an example for the GPT input format for clarity. GPT-3.5 is used for fine-tuning, and GPT-4 is used for both few-shot and zero-shot learning since GPT-4 is not yet available for fine-tuning as of May 2024.

**Model performance evaluations.** The performances of all models across various input formats and learning methods are illustrated in **Fig. 6**. MaTableGPT extracts data from tables and generates outputs in JSON format. Thus, a correct JSON output needs to have a hierarchical structure matching that of the keys, and the corresponding values must also be accurate. If the hierarchy of keys is different, even if the values are correctly extracted, the predicted output cannot be considered accurate. Therefore, model performance evaluations can be performed through two processes: (1) assessing whether the hierarchical structure of keys has been correctly extracted; this assessment is measured by the F1 score for the key structure (or "structure F1 score" hereafter), and (2) determining whether the values are correct when the key hierarchy is properly generated; this is evaluated through "value accuracy." Finally, the evaluation method further incorporates their harmonic mean, termed the "total F1 score." Detailed information on the evaluation method is provided in the methods section.

First, the structure F1 score is significantly affected by the input format. The baseline model includes extensive and unnecessary HTML tags and consistently performed worse than the same model with TSV representation. Throughout the paper, the model names are presented in the form of "input format (table splitting condition)," as shown by TSV (non-split). The superior key structure performance of TSV (non-split) over the baseline can be attributed to the reduction of input noise by minimizing the unnecessary HTML tags and the creation of the customized tags for data extraction from captions and titles. Our second strategy, table splitting, yielded even more pronounced improvements. In the case of TSV (split), the few-shot and fine-tuning models achieved structure F1 scores of 93.1% and 94.4%, respectively; these values exceeded the performance of TSV (non-split) with structure F1 scores of 79.3% and 89.0%, respectively. These results indicated that reducing input complexity by splitting tables was a highly effective approach. Moreover, compared with those of the baseline models, the structure F1 scores of JSON (non-split) were similar or slightly lower in both the fine-tuning and few-shot scenarios; these results could be attributed to the inherent errors in generating a column-based JSON structure, as discussed in the section on **table data representation and table splitting**. However, the improvement in the structure F1 scores through splitting demonstrated that splitting was a highly effective strategy regardless of the table data representation method used, thus validating its efficacy.

Next, for the analysis of the metric of value accuracy, both the few-shot and fine-tuning models achieved value accuracies exceeding 95%. However, the zero-shot model had relatively low accuracy mainly due to its ability to extract expressions directly from the table without normalization; this led to inconsistencies when the same data are represented differently across tables. Since the value accuracy is calculated only when the key structure is correct, a high value accuracy indicates that MaTableGPT's ability to accurately extract data depends on its proficiency in identifying the correct key structure.

Finally, the total F1 score is the harmonic mean of the structure F1 score and value accuracy and provides an overview of the overall performance of all models. The best-performing model employed both TSV representation with table splitting and an approach that eliminated hallucinated information through follow-up questions; this model achieved a performance of 96.8%.

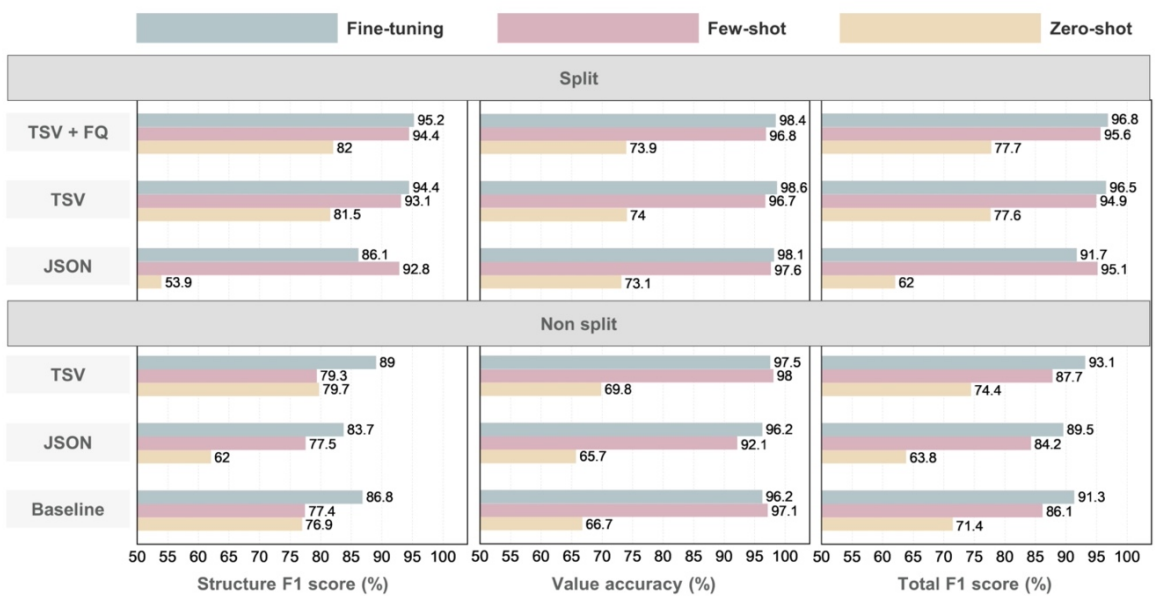

**Fig. 6 Performance comparison of the MaTableGPT models with various parameters of input formats, learning methods, and table splitting.** The input formats of customized TSV, customized JSON, and the baseline are considered. Baseline denotes the case of the original HTML format and

table non-splitting. The learning methods of fine-tuning, few-shot learning, and zero-shot learning are considered. Non-split refers to the table inputs that have not been split, while split denotes inputs where the table has been split. The metrics of the structure F1 score, value accuracy, and total F1 score are used.

**MaTableGPT accuracy-cost map.** Thus far, the extraction accuracy of MaTableGPT has been investigated. On the other hand, when using GPT for data extraction, two critical cost factors exist: labeling cost and LLM usage costs. Labeling requires a significant investment of time from skilled researchers, and achieving flawless and consistent labeling can be labor intensive due to the human nature of the task. The cost can increase further when multiple researchers are involved in labeling due to ambiguous standards. In addition, the cost associated with training and testing proprietary LLMs such as GPT is also substantial. For instance, GPT-4-1106-preview has a cost of $10.00 per 1M input tokens and $30.00 per 1M output tokens as of May 2024.

In this subsection, we examine the extraction accuracy, labeling cost, and GPT usage cost of all investigated MaTableGPT models, with the aim of identifying the most balanced solutions. **Fig. 7** shows the extraction accuracy and cost (both labeling cost and GPT cost) of different approaches based on the use of table data representation, application of table splitting, choice of learning methods, and application of follow-up questions. Note that for the labeling cost, zero-shot learning requires no labeling, few-shot learning requires 10 I/O paired labels, and the fine-tuning approach involves up to 126 I/O paired labels.

The Pareto-front solutions in terms of GPT cost and extraction accuracy are highlighted in **Fig. 7**. We first note that the highest-performing case is fine-tuning with follow-up questions and TSV (split); however, this approach results in more than ten times the labeling cost compared to few-shot method, and the GPT cost increases due to follow-up questions. This method could be ideal for those with sufficient human resources and budgets.

We propose two other MaTableGPT models on the Pareto-front line as the most balanced solutions: the 10-shot learning method with either TSV (split) or JSON (split). The 10-shot method exhibits slightly lower performance than the fine-tuning method. However, the significant reduction in the labeling requirements of the 10-shot learning method with respect to the fine-tuning method leads to substantial savings in labor and time costs; thus, using the 10-shot approach for data extraction is a very rational choice. In practice, both the TSV (split) and JSON (split) methods achieve performance levels of approximately 95%. Despite the tables being split, the GPT cost remains comparable to that of fine-tuning without table splitting. For those who aim to minimize the labeling costs and can afford a high GPT usage cost, few-shot learning with TSV (split) and improved performance through follow-up questions can be a viable option. As such, the accuracy-cost map in **Fig. 7** provides multiple solutions, enabling researchers to tailor their methods based on their specific circumstances and to balance labeling cost, GPT cost, and extraction accuracy.

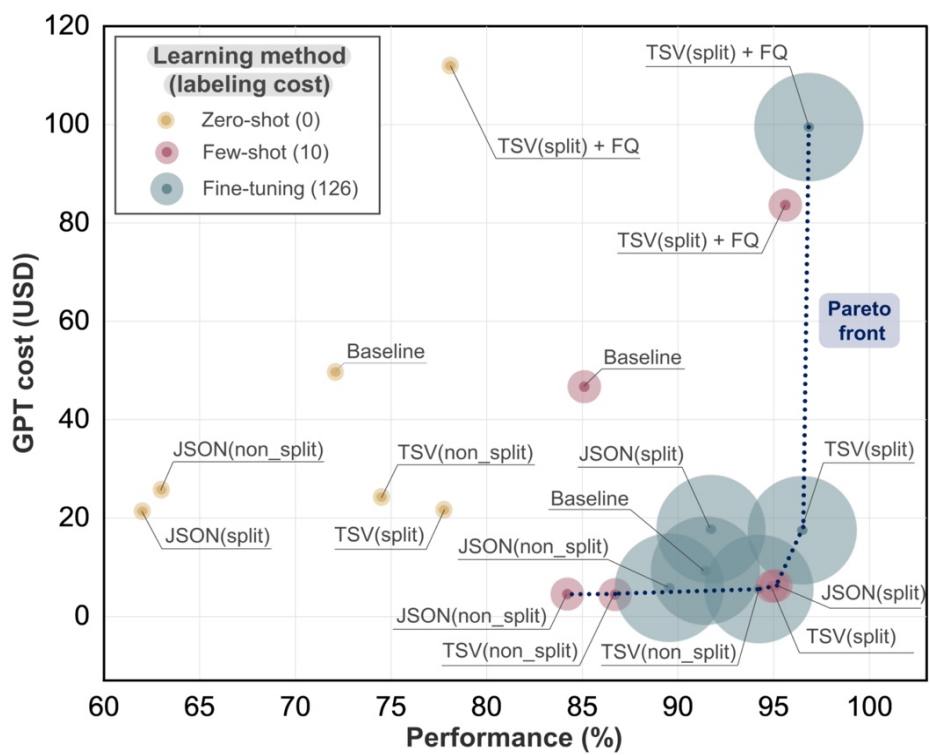

**Fig 7. MaTableGPT accuracy-cost map.** A map providing an overview of GPT usage cost, labeling cost, and performance (total F1 score) across the various input formats and models. For the labeling cost, the size of the circle represents the relative size of the labeling set, with larger circles indicating a larger labeling size. For fine-tuning, the GPT cost is calculated based on all tokens used in training and test inputs and outputs. For few-shot learning, the cost includes tokens from the task description, 10-shot examples, and their outputs. For zero-shot learning and follow-up questions, the cost includes tokens from the task description and outputs. For cases involving table splitting, the training set (used only for fine-tuning) contains 1,055 tables, while the test set contains 293 tables. For cases not involving table splitting, the training set for fine-tuning contains 126 tables, and the test set contains 35 tables. Details of each GPT cost can be found in Supplementary Note 3.

**Statistical analysis of the database on water splitting catalysis.** The application of the pretrained MaTableGPT to all prepared tables led to the construction of a large-scale database of water splitting catalysis, and several statistical data mining and analyses were performed, as shown in **Figs. 8** and **9**. Although OER catalysts were mainly targeted in the paper screening process, HER catalysts were also extracted because both the OER and HER catalysts were listed in the same table.

**Fig. 8a** shows the overpotential distributions of the OER and HER catalysts in different electrolyte environments (acidic vs. alkaline media). Here, the OER catalysts predominantly exhibited overpotentials within the range of 200-400 mV, whereas HER catalysts were primarily clustered around overpotentials of 50-150 mV. These results strongly support the well-known fact that OER reactions serve as a difficulty compared to HER reactions in terms of overpotential[30]. Additionally, catalysts with acidic electrolytes exhibited lower OER overpotentials than those with alkaline media, as confirmed by the left shift of the distribution curve. This statistical finding also agreed with the observations that catalysts involving Ir, Ru, and Pt, which are effective and commonly utilized in acidic media, demonstrated superior performance[31,32].

The database under examination consists of overpotential data gathered across a spectrum of current densities. An elevated current density causes intensified ion oscillations near the electrode interface, thereby perturbing the equilibrium of the electric double layer and consequentially increasing the overpotential. Moreover, the migration of ions toward the electrode during current passage leads to the formation of a boundary layer between the electrode and electrolyte, thereby increasing resistance and voltage dissipation. Consequently, the increase in the current density intensifies the resistance losses, leading to elevated overpotentials. **Fig. 8b** shows insightful comparisons delineating the overpotential distributions for two distinct current densities sourced from the database. **Fig. 8b** shows the overpotential distributions corresponding to 10 mA/cm² and 100 mA/cm² for both the OER and HER processes. Notably, across both reactions, heightened overpotentials are evident at a relatively elevated current density of 100 mA/cm².

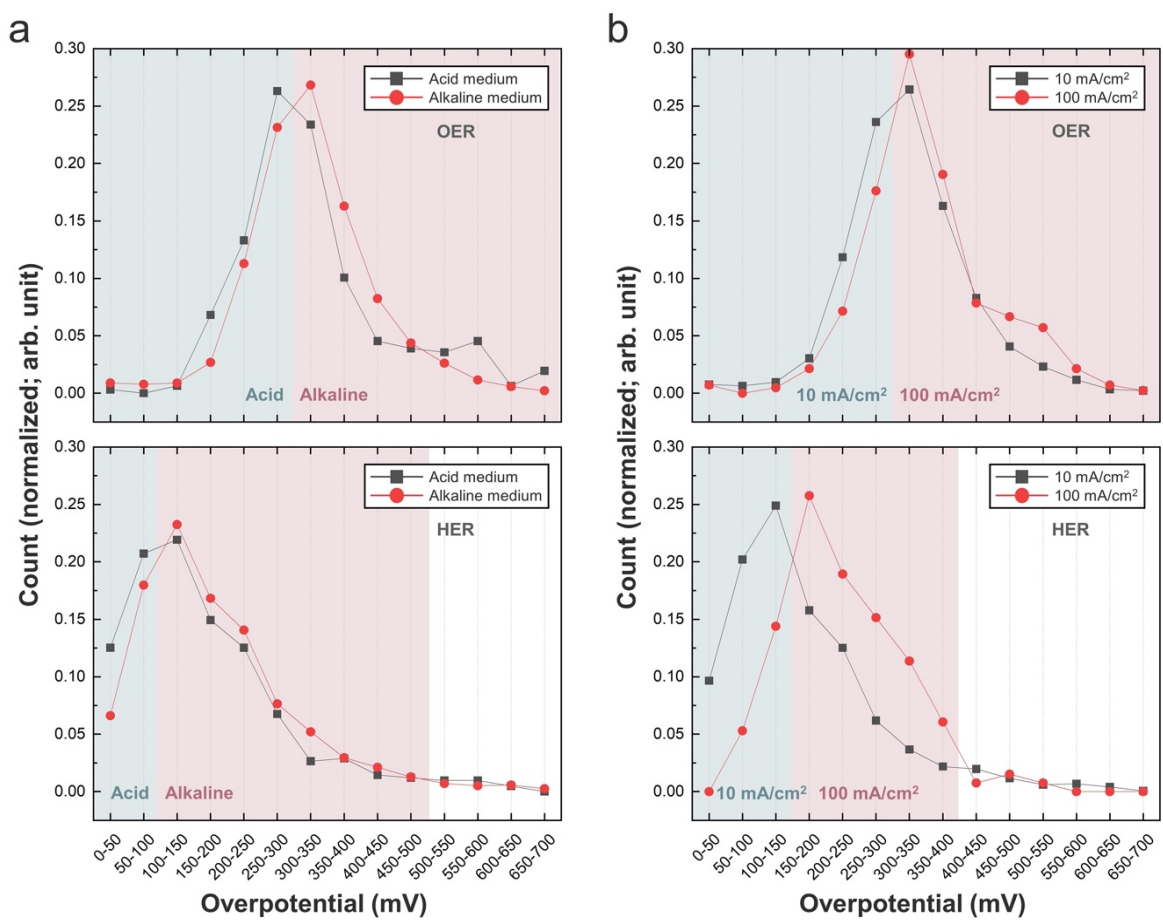

**Fig. 8 Distribution of the overpotentials under different electrolyte conditions and different current densities.** (**a**) Distribution of the overpotentials for each OER and HER under different electrolyte conditions (acidic and alkaline). (**b**) Distribution of the overpotentials for each OER and HER at different current densities (10 mA/cm² and 100 mA/cm²). Y-axis represents count normalized by dividing by the total sum of counts.

Exploring the vast catalyst database reveals intriguing insights into element utilization across different electrolyte environments. **Fig. 9a** illustrates the element occurrence frequencies in each acidic and alkaline medium, highlighting their prevalence in distinct electrochemical settings. Next, **Figs. 9b-9g** show the association rule mining (ARM) to reveal concurrent relationships among elements, with node sizes indicating the occurrence frequencies and edge thickness representing co-usage frequency. This analysis reveals intricate associations in element utilization, providing valuable insights for catalytic design and optimization.

In acidic environments with relatively low pH, the generation of hydrogen at the cathode during the electrochemical decomposition of water shows excellent kinetics with Pt-based catalysts[33,34]. However, finding stable OER catalysts at the anode in harsh oxidative environments with low pH values and high electrode potentials remains a major challenge. Thus far, Ru has been generally recognized as a material with the highest catalytic activity, and numerous studies have used Ru and Ir-based catalysts because $IrO_2$ catalysts exhibit greater corrosion stability than Ru-based catalysts[35,36]. **Fig. 9a** confirms the widespread use of Ru and Ir, with Ru-based catalysts generally exhibiting lower overpotentials than Ir-based catalysts. Additionally, transition metals such as Ni, Co, and Fe have been extensively utilized with Ir and Ru[37,38]. Generally, the formation of doped catalysts is considered an effective method for enhancing the activity and stability of catalysts by adjusting their electronic structures[39,40]. Therefore, elements such as Ni, Co, and Sr are likely doped into $RuO_2$- or $IrO_2$-based catalysts.

Indeed, in **Fig. 9c**, when the elements used with Ir are examined, Ni, Fe, Co, and other doping elements in addition to Ru are observed. Notably, Sr is frequently used alongside Ir; these results reflect the research involving $IrO_x/SrIrO_3$ catalysts, which have been reported to exhibit high OER performance[41,42], and subsequent studies in which $SrIrO_3$ perovskite was doped with various elements, such as Co and Ti[43,44]. **Fig. 9d** shows the elements used alongside Ru, with Ir being the most commonly used in combination with Ru. This reflects the efforts of combining $IrO_2$ with $RuO_2$ to improve the catalytic stability.

Alkaline environments provide enhanced corrosion durability due to the relatively high pH, and the utilization of a variety of transition metals as electrode materials becomes feasible; thus, expensive noble metals such as Ir and Ru are replaced. Metals such as Ni, Co, and Fe[45,46], as illustrated in **Fig. 9a**, are commonly employed in these scenarios. First-row transition metals such as Fe, Co, and Ni exhibit diverse oxidation states and electron exchange capabilities; therefore, they are considered to be highly active catalysts.

**Figs. 9f** and **9g** show the frequent co-usage of Ni and Co with other elements, respectively. Ni has demonstrated excellent performance, particularly in studies employing NiFe LDH[47,48]; these results are consistent with the thickest edge between Ni and Fe shown in **Fig. 9e**. Regarding oxides, $Fe_2O_3$ and $NiO$ are most stable in single $Fe^{3+}$ and $Ni^{2+}$ states, respectively; however, $Co_3O_4$ consists of $Co^{2+}$ and $Co^{3+}$ states and can undergo changes under oxidation–reduction conditions[49,50,51]. Due to this characteristic, Co is ideal for forming various compounds; thus, it is extensively used as much as Ni. Conversely, although Fe demonstrates comparatively lower activity in contrast to Ni and Co-based catalysts, its multifaceted electronic structures play an instrumental role in strengthening stability[52]. Consequently, Fe is commonly employed in conjunction with Ni or Co, as illustrated in **Figs. 9f** and **9g**.

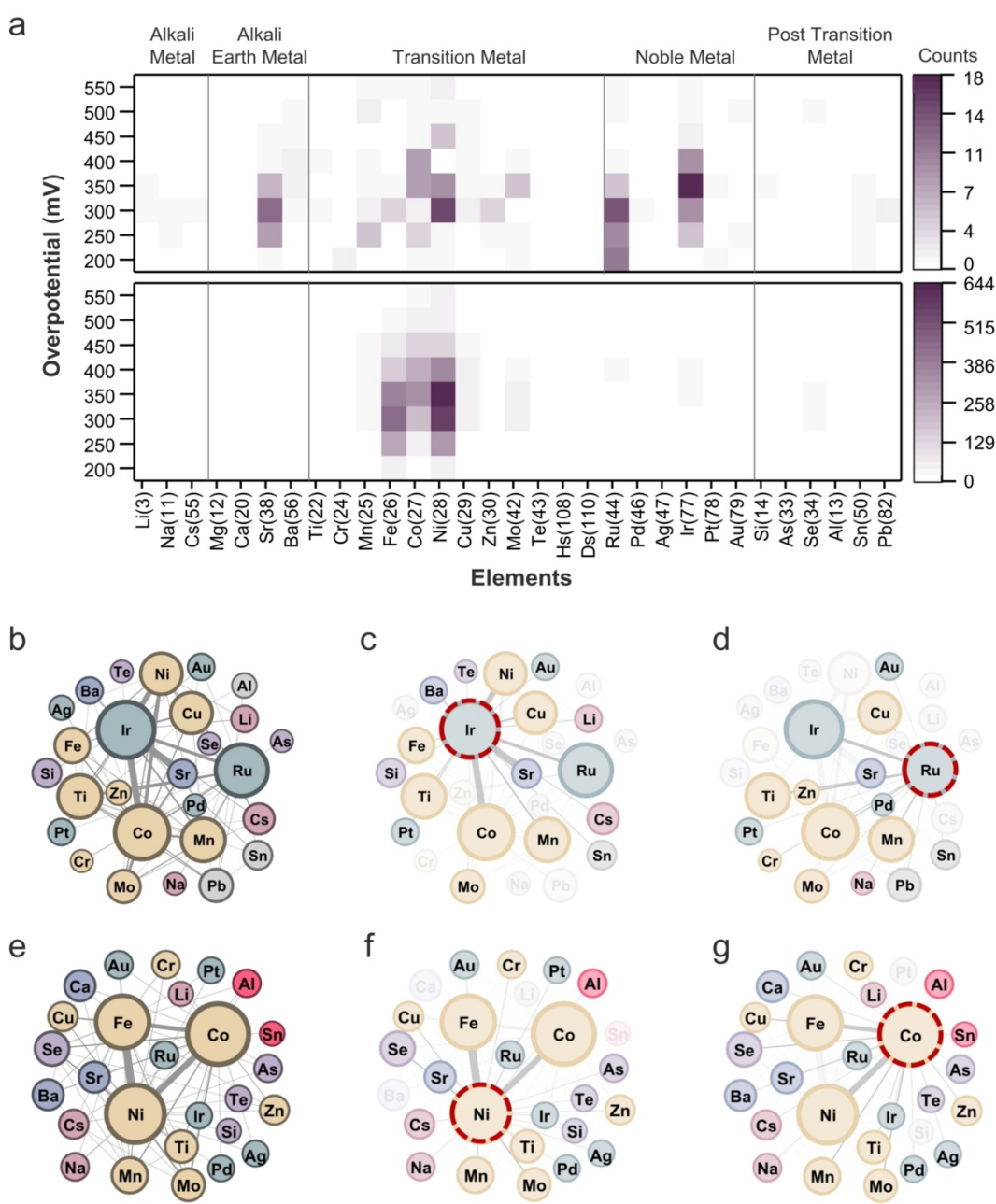

**Fig. 9 Elemental utilization across different electrolyte environments.** (**a**) Heatmap of the most commonly used elements in OER catalysts in each acidic and alkaline medium. (**b-g**) Visualization of the ARM results. (**b**) Entire graph of the ARM results for an acidic medium. (**d**) Subgraph highlighting the nodes that are connected to Ir for acidic media. (**d**) Subgraph highlighting the nodes that are connected to Ru for acidic media. (**e**) Entire graph of the ARM results for alkaline media. (**f**) Subgraph highlighting the nodes that are connected to Ni for alkaline media. (**g**) Subgraph highlighting the nodes that are connected to Co for alkaline media.

## Discussion

In the field of material science, extracting table data remains an unexplored area. In this study, based on the development of MaTableGPT, highly accurate extraction of table data could be achieved, thus providing a high-quality materials database. We showed exceptional performance, achieving a total F1 score of up to 96.8% through combined strategies of TSV

(split) input processing, GPT fine-tuning, and follow-up questions. Furthermore, our objective was to pinpoint the most balanced and optimal solutions by assessing the extraction accuracy, labeling expenses, and GPT utilization costs across all evaluated MaTableGPT models. Consequently, we proposed the few-shot learning approach based on TSV (split) processing as the best solution, which achieved a nearly 95% total F1 score, with only 10 I/O labeling costs and minimal GPT usage costs (5.9 US dollars); thus, it was the most reasonable solution for MaTableGPT. Exploiting the advantages of low labeling costs in few-shot learning, MaTableGPT could efficiently extract table data across various fields, such as catalysis and batteries, regardless of domain.

Using the pretrained MaTableGPT model (fine-tuned model based on TSV (split)), we constructed a database related to the performance of OER and HER catalysts; this database contained information on catalysts, their performance, and various performance attributes. The statistical analyses conducted on the MaTableGPT-extracted database provided valuable insights into the distribution of the overpotentials and elemental utilization of reported catalysts for water splitting. This demonstration showed the significant advantage of MaTableGPT since it enabled the accurate and cost-competitive extraction of a vast amount of information existing in tables in the materials science literature via a single process and, as a result, enabled insightful and statistical analysis.

# Method

**Content acquisition.** The electrochemical water splitting reactions involve both the OER and HER, and these reactions occur at the anode and cathode, respectively. Understanding that the OER is regarded as a difficulty in water splitting, we intentionally used the OER as a keyword for paper screening to exclude papers without involving the OER. The papers were collected from six accredited publishers: American Chemical Society (ACS), Elsevier, Royal Society of Chemistry (RSC), American Association for the Advancement of Science (AAAS), Nature Publishing Group (NPG), and Wiley. The search began with keywords such as 'OER electrocatalyst water splitting' or 'OER electrocatalyst water oxidation'; this led to the initial discovery of 26,330 papers. To refine the search, the papers containing keywords such as 'battery,' '$CO_2$,' 'fuel cell,' 'methanol,' and '$H_2O_2$' in their titles were excluded; this process filtered out the noise, and 22,561 papers remained. Subsequent analysis involved term frequency-inverse document frequency (TF-IDF)[53], which is a statistical measure assessing the relevance of a term within a document and facilitating the classification of similar documents from a vast dataset. TF-IDF values were calculated for all filtered papers using the entire main body of each document. The papers were then excluded if their TF-IDF value for 'OER' fell below those for 'battery,' '$CO_2$,' 'aldehyde,' 'alcohol,' 'ORR,' and 'photo'; this process provides a collection of 11,077 papers in the water splitting catalysis domain.

**Output structure design.** A table is a structured format, yet the diversity within its contents is vast. For instance, while a performance table in the OER literature is assumed to contain only information on the OER catalysts, in reality, it often contains various catalysts for reactions such as the OER, HER, and bifunctional catalysts, among others. Thus, crafting a well-structured JSON format for extracting table data is crucial. The architecture of our constructed database is hierarchical, organized as catalyst-performance-property (performance attributes). Properties include 'electrolyte,' 'reaction_type,' 'value,' 'current_density,' 'overpotential,' 'potential,' 'substrate,' and 'versus.' Information such as 'reaction_type' and 'substrate' are potentially considered common across the entire table; however, in practice, numerous cases exist where this assumption does not hold; hence, this information is placed at the lowest level of hierarchy, which is at the property level.

**Rule-based sub-header recognition.** For table splitting, specific rules were applied to identify the presence of headers within the HTML body. Since the target was tables containing catalyst performance data, numerical values were assumed to be included in the performance metrics. If there was not at least one cell in a row within the HTML body whose first character, excluding special characters (e.g., ~, <, >), was a number, then, that row was considered a sub-header. Additionally, if all cells in a row within the body were merged into one cell, it was also considered a sub-header. These assumptions effectively filtered out tables containing sub-headers.

**GPT details.** We utilized GPT-3.5-turbo-1106 for fine-tuning, GPT-4-1106-preview for zero-shot and few-shot learning, and GPT-4-0125-preview for follow-up questions. These were selected because GPT-4 is not available for fine-tuning or the service (fine-tuning in GPT-4) is limited to selected groups as of May 2024. All models were trained with the following set parameters: temperature = 0, frequency_penalty = 0, and presence_penalty = 0.

**Model performance evaluation method.** Evaluation of the GPT output that generates JSON data is a crucial process because it ensures the fidelity of the generated contents. When the JSON structures are created, the accuracy of the extraction of the hierarchy of keys and their relationships need to be assessed. Additionally, an evaluation of the correctness of the values extracted is equally vital.

First, for key evaluation, the JSON structure typically maintains a consistent hierarchical arrangement of keys concerning the catalyst name, reaction type, catalytic performance, and measurement conditions. Therefore, any discrepancies in the hierarchical key arrangement could lead to inaccuracies, even if the values are correct. Given the propensity for varied erroneous outcomes in generative models, encompassing these diverse errors into evaluation metrics entails employing true positives (TPs), false negatives (FNs), and false positives (FPs) to compute the "structure F1 score" as follows.

$$\text{Structure F1 score} = \frac{TP}{TP + 1/2(FN + FP)}$$

If the key was generated correctly, it was considered a true positive; if a key was not generated that should have been, it was considered a false negative; and if a key was generated that should not have been, it was considered a false positive.

On the other hand, for the value evaluation, the focus is exclusively on instances where the key structure is accurately replicated. The "value accuracy" is computed as follows:

$$\text{Value accuracy} = \frac{\text{\# of correct predictions}}{\text{\# of correct predictions} + \text{\# of incorrect predictions}}$$

Finally, the evaluation method further incorporates their harmonic mean, termed the "total F1 score." The evaluation approach for each key and value ensures that not only the hierarchical arrangement of keys but also the correctness of the values extracted are thoroughly assessed. Discrepancies such as those involving spacing, singular/plural forms, and similar nuances, albeit present in the correct answer, are still categorized as incorrect within this meticulous evaluation process. A detailed example of the evaluation method can be found in Supplementary Note 4.

**Prompt engineering.** The performance of the GPT model is widely known to be influenced by the prompts used[54,55,56]. According to Zheng *et al*., prompts that provide detailed instructions, request structured output, and minimize hallucinations can enhance the performance of the GPT 3.5 model[57]. To this end, prompt engineering was conducted to enhance the performance of our method. The tested prompts included a simple description of the task (Prompt #1 with a total F1 score of 0.930), a task description with an added list of the features to be extracted (Prompt #2 with a total F1 score of 0.937), and a task description with a list of features to be extracted and example outputs (Prompt #3 with a total F1 score of 0.968). All GPT fine-tuning results were produced based on Prompt #3. Detailed descriptions of these prompt engineering procedures can be found in Supplementary Note 5.

## Data availability

The performance data of OER catalysts obtained through MaTableGPT can be found at https://zenodo.org/doi/10.5281/zenodo.11362347. Each dataset consists of the DOI of the paper containing the table, the name of the catalyst, and the performance and properties of the catalyst.

## Code Availability

All table-mining codes and related information in this work are available at https://github.com/KIST-CSRC/MaTableGPT or can be obtained from corresponding authors upon request.

## Acknowledgements
This work was supported by the National Center for Materials Research Data (NCMRD) through the National Research Foundation of Korea funded by the Ministry of Science and ICT (NRF-2022M3H4A7046278 and NRF2021M3A7C2089739) and KIST institutional projects (2E33211 and 2Z07160).

## Author contributions
D. K. conceived the idea. A. H., S. S. H. and D. K. supervised the project. G. H. Y. and Jiwoo Choi developed the entire protocols of MaTableGPT and performed all ML training and test. G. H. Y., Jiwoo Choi. and H. S. contributed to the developments of input processing within MaTableGPT. O. M., Jaewoong Choi, K. B., B. L., D. B., A. H. and S. S. H. contributed to the analysis of table data extraction results. G. H. Y., Jiwoo Choi and D. K. wrote the manuscript with the inputs from all authors.

## Competing interests
The authors declare no competing interest

# References

1.  Himanen, L., Geurts, A., Foster, A. S. & Rinke, P. Data-driven materials science: status, challenges, and perspectives. *Adv. Sci.* **6**, 1900808 (2019).

2.  Cole, J. M. A design-to-device pipeline for data-driven materials discovery. *Acc. Chem. Res.* **53**, 599–610 (2020).

3.  Ramprasad, R., Batra, R., Pilania, G., Mannodi-Kanakkithodi, A. & Kim, C. Machine learning in materials informatics: recent applications and prospects. *npj Comput. Mater.* **3**, 54 (2017).

4.  Jain, A., Hautier, G., Ong, S. P. & Persson, K. New opportunities for materials informatics: resources and data mining techniques for uncovering hidden relationships. *J. Mater. Res.* **31**, 977–994 (2016).

5.  Jain, A. *et al.* Commentary: The Materials Project: A materials genome approach to accelerating materials innovation. *APL Mater.* **1**, (2013).

6.  Saal, J. E., Kirklin, S., Aykol, M., Meredig, B. & Wolverton, C. Materials design and discovery with high-throughput density functional theory: the open quantum materials database (OQMD). *Jom* **65**, 1501–1509 (2013).

7.  Draxl, C. & Scheffler, M. NOMAD: The FAIR concept for big data-driven materials science. *Mrs Bull.* **43**, 676–682 (2018).

8.  Tran, R. *et al.* The Open Catalyst 2022 (OC22) dataset and challenges for oxide electrocatalysts. *ACS Catal.* **13**, 3066–3084 (2023).

9.  Swain, M. C. & Cole, J. M. ChemDataExtractor: a toolkit for automated extraction of chemical information from the scientific literature. *J. Chem. Inf. Model.* **56**, 1894–1904 (2016).

10. Mavracic, J., Court, C. J., Isazawa, T., Elliott, S. R. & Cole, J. M. ChemDataExtractor 2.0: Autopopulated ontologies for materials science. *J. Chem. Inf. Model.* **61**, 4280–4289 (2021).

11. Wang, L. *et al.* A corpus of CO2 electrocatalytic reduction process extracted from the scientific literature. *Sci. Data* **10**, 175 (2023).

12. Choi, J. *et al.* Deep learning of electrochemical CO 2 conversion literature reveals research trends and directions. *J. Mater. Chem. A* (2023).

13. Mukaddem, K. T., Beard, E. J., Yildirim, B. & Cole, J. M. ImageDataExtractor: a tool to extract and quantify data from microscopy images. *J. Chem. Inf. Model.* **60**, 2492–2509 (2019).

14. Luo, J., Li, Z., Wang, J. & Lin, C.-Y. Chartocr: Data extraction from charts images via a deep hybrid framework. in *Proceedings of the IEEE/CVF winter conference on applications of computer vision* 1917–1925 (2021).

15. Kato, H., Nakazawa, M., Yang, H.-K., Chen, M. & Stenger, B. Parsing line chart images using linear programming. in *Proceedings of the IEEE/CVF Winter Conference on Applications of Computer Vision* 2109–2118 (2022).

16. Hassan, M. Y. & Singh, M. LineEX: Data Extraction From Scientific Line Charts. in *Proceedings of the IEEE/CVF Winter Conference on Applications of Computer Vision* 6213–6221 (2023).

17. Dagdelen, J. *et al.* Structured information extraction from scientific text with large language models. *Nat. Commun.* **15**, 1418 (2024).

18. Brown, T. *et al.* Language models are few-shot learners. *Adv. Neural Inf. Process. Syst.* **33**, 1877–1901 (2020).

19. Achiam, J. *et al.* Gpt-4 technical report. *arXiv Prepr. arXiv2303.08774* (2023).

20. Choi, J. & Lee, B. Accelerating materials language processing with large language models. *Commun. Mater.* **5**, 13 (2024).

21. Xie, T. *et al.* Large language models as master key: unlocking the secrets of materials science with GPT. *arXiv Prepr. arXiv2304.02213* (2023).

22. Kang, Y. & Kim, J. Chatmof: An autonomous ai system for predicting and generating metal-organic frameworks. *arXiv Prepr. arXiv2308.01423* (2023).

23. Lee, W., Kang, Y., Bae, T. & Kim, J. Harnessing Large Language Model to collect and analyze Metal-organic framework property dataset. *arXiv Prepr. arXiv2404.13053* (2024).

24. Hanan, A. *et al.* An efficient and durable bifunctional electrocatalyst based on PdO and Co2FeO4 for HER and OER. *Int. J. Hydrogen Energy* **48**, 19494–19508 (2023).

25. Wang, J. *et al.* Self-standing and efficient bifunctional electrocatalyst for overall water splitting under alkaline media enabled by Mo1-xCoxS2 nanosheets anchored on carbon fiber paper. *Int. J. Hydrogen Energy* **44**, 13205–13213 (2019).

26. Wang, Z.-J., Jin, M.-X., Zhang, L., Wang, A.-J. & Feng, J.-J. Amorphous 3D pomegranate-like NiCoFe nanoassemblies derived by bi-component cyanogel reduction for outstanding oxygen evolution reaction. *J. Energy Chem.* **53**, 260–267 (2021).

27. Yu, A., Kim, M. H., Lee, C. & Lee, Y. Structural transformation between rutile and spinel crystal lattices in Ru–Co binary oxide nanotubes: enhanced electron transfer kinetics for the oxygen evolution reaction. *Nanoscale* **13**, 13776–13785 (2021).

28. Zhang, T., Wu, F., Katiyar, A., Weinberger, K. Q. & Artzi, Y. Revisiting few-sample BERT fine-tuning. *arXiv Prepr. arXiv2006.05987* (2020).

29. Sun, C., Qiu, X., Xu, Y. & Huang, X. How to fine-tune bert for text classification? in *Chinese computational linguistics: 18th China national conference, CCL 2019, Kunming, China, October 18–20, 2019, proceedings 18* 194–206 (Springer, 2019).

30. Sun, H. *et al.* Electrochemical water splitting: Bridging the gaps between fundamental research and industrial applications. *Energy Environ. Mater.* **6**, e12441 (2023).

31. An, L. *et al.* Recent development of oxygen evolution electrocatalysts in acidic environment. *Adv. Mater.* **33**, 2006328 (2021).

32. Hansen, J. N. *et al.* Is there anything better than Pt for HER? *ACS energy Lett.* **6**, 1175–1180 (2021).

33. Zhao, G., Rui, K., Dou, S. X. & Sun, W. Heterostructures for electrochemical hydrogen evolution reaction: a review. *Adv. Funct. Mater.* **28**, 1803291 (2018).

34. Li, C. & Baek, J.-B. Recent advances in noble metal (Pt, Ru, and Ir)-based electrocatalysts for efficient hydrogen evolution reaction. *ACS omega* **5**, 31–40 (2019).

35. Reier, T., Oezaslan, M. & Strasser, P. Electrocatalytic oxygen evolution reaction (OER) on Ru, Ir, and Pt catalysts: a comparative study of nanoparticles and bulk materials. *Acs Catal.* **2**, 1765–1772 (2012).

36. Spöri, C. *et al.* Experimental activity descriptors for iridium-based catalysts for the electrochemical oxygen evolution reaction (OER). *ACS Catal.* **9**, 6653–6663 (2019).

37. Li, P. *et al.* Boosting oxygen evolution of single-atomic ruthenium through electronic coupling with cobalt-iron layered double hydroxides. *Nat. Commun.* **10**, 1711 (2019).

38. Wu, Z.-Y. *et al.* Non-iridium-based electrocatalyst for durable acidic oxygen evolution reaction in proton exchange membrane water electrolysis. *Nat. Mater.* **22**, 100–108 (2023).

39. Feng, Y.-Y. *et al.* Copper-doped ruthenium oxide as highly efficient electrocatalysts for the evolution of oxygen in acidic media. *J. Alloys Compd.* **892**, 162113 (2022).

40. Zoller, F. *et al.* Carbonaceous Oxygen Evolution Reaction Catalysts: From Defect and Doping-Induced Activity over Hybrid Compounds to Ordered Framework Structures. *Small* **17**, 2007484 (2021).

41. Wan, G. *et al.* Amorphization mechanism of SrIrO3 electrocatalyst: How oxygen redox initiates ionic diffusion and structural reorganization. *Sci. Adv.* **7**, eabc7323 (2021).

42. Seitz, L. C. *et al.* A highly active and stable IrO x/SrIrO3 catalyst for the oxygen evolution reaction. *Science (80-. ).* **353**, 1011–1014 (2016).

43. Liang, X. *et al.* Activating inert, nonprecious perovskites with iridium dopants for efficient oxygen evolution reaction under acidic conditions. *Angew. Chemie Int. Ed.* **58**, 7631–7635 (2019).

44. Zhao, J.-W. *et al.* The formation of unsaturated IrOx in SrIrO3 by cobalt-doping for acidic oxygen evolution reaction. *Nat. Commun.* **15**, 2928 (2024).

45. Xie, X. *et al.* Oxygen evolution reaction in alkaline environment: material challenges and solutions. *Adv. Funct. Mater.* **32**, 2110036 (2022).

46. Mohammed-Ibrahim, J. A review on NiFe-based electrocatalysts for efficient alkaline oxygen evolution reaction. *J. Power Sources* **448**, 227375 (2020).

47. Bodhankar, P. M., Sarawade, P. B., Singh, G., Vinu, A. & Dhawale, D. S. Recent advances in highly active nanostructured NiFe LDH catalyst for electrochemical water splitting. *J. Mater. Chem. A* **9**, 3180–3208 (2021).

48. Alobaid, A., Wang, C. & Adomaitis, R. A. Mechanism and Kinetics of HER and OER on NiFe LDH Films in an Alkaline Electrolyte. *J. Electrochem. Soc.* **165**, J3395–J3404 (2018).

49. Smith, A. L., Hardcastle, K. I. & Soper, J. D. Redox-active ligand-mediated oxidative addition and reductive elimination at square planar cobalt (III): Multielectron reactions for cross-coupling. *J. Am. Chem. Soc.* **132**, 14358–14360 (2010).

50. Popat, Y. *et al.* Pulsed laser deposition of CoFe2O4/CoO hierarchical-type nanostructured heterojuction forming a Z-scheme for efficient spatial separation of photoinduced electron-hole pairs and highly active surface area. *Appl. Surf. Sci.* **489**, 584–594 (2019).

51. Zhang, R. *et al.* Engineering cobalt defects in cobalt oxide for highly efficient electrocatalytic oxygen evolution. *Acs Catal.* **8**, 3803–3811 (2018).

52. Anantharaj, S., Kundu, S. & Noda, S. "The Fe Effect": A review unveiling the critical roles of Fe in enhancing OER activity of Ni and Co based catalysts. *Nano Energy* **80**, 105514 (2021).

53. Joachims, T. A probabilistic analysis of the Rocchio algorithm with TFIDF for text categorization. in *ICML* vol. 97 143–151 (Citeseer, 1997).

54. Kojima, T., Gu, S. S., Reid, M., Matsuo, Y. & Iwasawa, Y. Large language models are zero-shot reasoners. *Adv. Neural Inf. Process. Syst.* **35**, 22199–22213 (2022).

55. Polak, M. P. & Morgan, D. Extracting accurate materials data from research papers with conversational language models and prompt engineering. *Nat. Commun.* **15**, 1569 (2024).

56. Amatriain, X. Prompt design and engineering: Introduction and advanced methods. *arXiv Prepr. arXiv2401.14423* (2024).

57. Zheng, Z., Zhang, O., Borgs, C., Chayes, J. T. & Yaghi, O. M. ChatGPT Chemistry Assistant for Text Mining and Prediction of MOF Synthesis. *arXiv Prepr. arXiv2306.11296* (2023).